\documentclass[runningheads,a4paper]{llncs}

\title{Simulation leagues: Enabling replicable and robust investigation of complex robotic systems}
\author{David M. Budden$^{1}$, Peter Wang$^2$, Oliver Obst$^2$, Mikhail Prokopenko$^3$}

\institute{$^1$ Systems Biology Laboratory, Melbourne School of Engineering, The University of Melbourne, VIC 3010, Australia \\
$^2$ Data Mining Team, CSIRO Digital Productivity, Epping, NSW 1710, Australia\\
$^3$ Complex Systems Research Group, Faculty of Engineering and IT, The University of Sydney, NSW 2006, Australia\\
        \email{david.budden@unimelb.edu.au}\\
        \email{\{peter.wang,oliver.obst\}@csiro.au}\\
				\email{mikhail.prokopenko@sydney.edu.au}}

\date{}

\usepackage[table]{xcolor}
\usepackage{times}
\usepackage{url}

\usepackage{graphicx}
\usepackage{amsmath}

\usepackage{epstopdf}

\begin{document}
\maketitle

\begin{abstract}
Physically-realistic simulated environments are powerful platforms for enabling measurable, replicable and statistically-robust investigation of complex robotic systems. Such environments are epitomised by the RoboCup simulation leagues, which have been successfully utilised to conduct massively-parallel experiments in topics including: optimisation of bipedal locomotion, self-localisation from noisy perception data and planning complex multi-agent strategies without direct agent-to-agent communication. Many of these systems are later transferred to physical robots, making the simulation leagues invaluable well-beyond the scope of simulated soccer matches.
In this study, we provide an overview of the RoboCup simulation leagues and describe their properties as they pertain to replicable and robust robotics research. To demonstrate their utility directly, we leverage the ability to run parallelised experiments to evaluate different competition formats (\emph{e.g.} round robin) for the RoboCup 2D simulation league. Our results demonstrate that a previously-proposed hybrid format minimises fluctuations from `true' (statistically-significant) team performance rankings within the time constraints of the RoboCup world finals.
Our experimental analysis would be impossible with physical robots alone, and we encourage other researchers to explore the potential for enriching their experimental pipelines with simulated components, both to minimise experimental costs and enable others to replicate and expand upon their results in a hardware-independent manner.
\end{abstract}

\section{Introduction}

Robotics researchers face many unique challenges when designing measurable, replicable and statistically-robust experiments. Robots are generally expensive and require regular maintenance, pressuring researchers to minimise the time spent evaluating algorithms and behaviours. This problem is amplified by the experimental confounds introduced by robots (\emph{e.g.} motor temperature) and their environments (\emph{e.g.} lighting variation), requiring many more experimental iterations to yield statistically-robust results. Moreover, the cost of physical robots makes platform-specific research replication particularly difficult, threatening the reliability of the peer review process in the presumably-common scenario of reviewers not having access to the robot-in-question.

The RoboCup simulation leagues and similar platforms provide a powerful framework for enabling replicable and robust investigation of complex robotic systems. Although these platforms are unable to perfectly model physical sources of system stochasticity, this shortcoming is greatly outweighed by their ability to abstract away the requirement for physical robots and inherent support for massively-parallel processing. In scenarios where imperfect system modelling is a major limitation, hybrid solutions have been developed that use simulated environments for initial experimentation (\emph{e.g.} global exploration of high-dimensional parameter spaces) and physical robots for fine-tuning (\emph{e.g.} local optimisation across principle components). The latter approach is epitomised by bipedal gait optimisation and has been shown to be highly effective for the Aldebaran NAO humanoid robot~\cite{macalpine2012design,niehaus2007gait}.

In this study, we provide an overview of the RoboCup simulation leagues in the context of enabling replicable and statistically-robust robotics research. We demonstrate the utility of massively-parallel experimentation by evaluating different competition formats (\emph{e.g.} round robin) for the RoboCup 2D simulation league. Tens-of-thousands of games were necessary to resolve non-transitivity and inherent stochastically of team performance, which would be intractable for 10-minute matches with physical robots. Our results demonstrate that a hybrid format best captures true team performance in the time constraints of the RoboCup world finals, and this format was subsequently adopted for the competitions at RoboCup 2014 Brazil.

\section{RoboCup simulation leagues}

RoboCup (the ``World Cup" of robot soccer) was first proposed in 1997 as a standard problem for the evaluation of theories, algorithms and architectures in areas including artificial intelligence (AI), robotics and computer vision~\cite{kitano1997robocup1}. This proposal followed the observation that traditional AI problems were increasingly unable to meet these requirements and that a new challenge was necessary to initiate the development of next-generation technologies.

The overarching RoboCup goal of developing a team of humanoid robots capable of defeating the FIFA World Cup champion team, coined the ``Millennium Challenge", has proven a major factor in driving research in AI and related areas for nearly two decades, with a search for the term ``RoboCup" in a major literature database yielding over 25,000 results. Since 1997, researchers and competitors have decomposed this ambitious pursuit into two complementary categories~\cite{kitano1997robocup1}:
\begin{itemize}
\item\textbf{Physical robot league:} Using physical robots to play soccer games. This category now contains many different leagues for both wheeled robots (small-sized and mid-sized leagues) and humanoids (standard platform and humanoid leagues), with each focusing on different aspects of physical robot design, motor control and bipedal locomotion, real-time localisation and computer vision~\cite{budden2013salient,fountain2013motivated}.
\item\textbf{Software agent league:} Using software or synthetic agents to play soccer games on an official soccer server over a network. This category contains both 2D~\cite{chen7robocup,gliders2013tdp,gliders2014tdp} and 3D~\cite{sim3d2013} simulation leagues.
\end{itemize}

The RoboCup simulation leagues traditionally involve the largest number of international participating teams, reaching 40 in 2013~\cite{bai2012wrighteagle}. The ability to simulate soccer matches without physical robots abstracts away low-level hardware and environmental issues (\emph{e.g.} motor temperature and breakages), allowing teams to focus on the development of complex team behaviours and strategies for a larger number of autonomous agents. Moreover, the simulation leagues often serve as platforms for the initial development and evaluation of software modules for later integration into physical robots~\cite{macalpine2012design,niehaus2007gait}. Many of these modules have applications well-beyond the RoboCup domain (\emph{e.g.} localisation and mapping~\cite{budden2013particle}), and the hardware-independent results inherent to simulated robots promote extension and replication by other researchers.

\subsection{Properties and utility of 2D and 3D leagues}

\begin{figure}[t]
\begin{center}
\includegraphics[width=15cm]{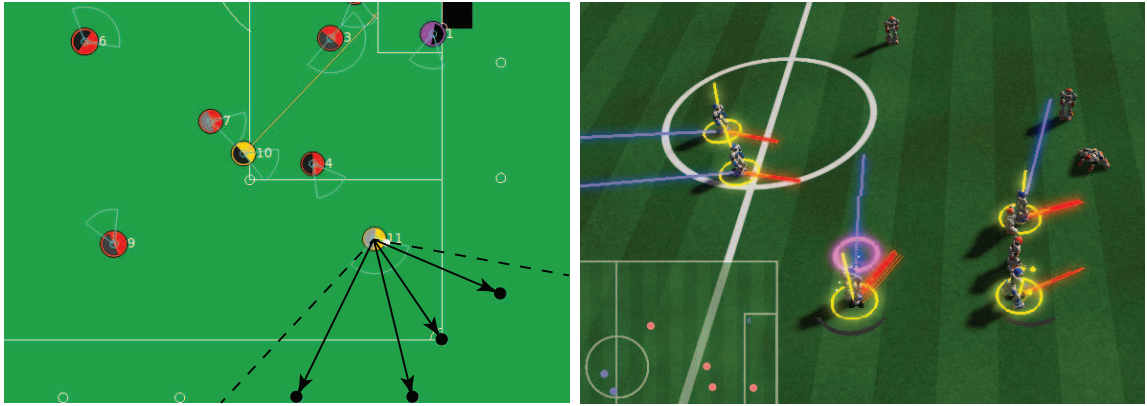}
\end{center}
\caption{Screenshots from the RoboCup 2D (left) and 3D (right) simulation leagues. The 2D simulation league screenshot demonstrates players from two teams (represented by red and yellow circles) and their respective fields-of-view. The black arrow overlays for player 11 illustrate how an agent can self-localise from observations of unique landmark features~\cite{budden2013particle}. The 3D simulation league screenshot demonstrates similar tactical overlays for each player. \label{fig:screenshotsim}}
\end{figure}

The RoboCup simulation league consists of both 2D and 3D competitions, which exhibit many similarities~\cite{bai2012wrighteagle}:

\begin{itemize}
\item The world model, including player and ball dynamics and kinematics, are simulated by a central \emph{soccer server}~\cite{chen7robocup}.
\item Participants develop a team of fully autonomous \emph{agents}, each of which interacts with the soccer server:
    \begin{itemize}
    \item Each agent receives information from the server regarding its current field of view.
    \item Each agent determines what actions to execute, and submits these requests to the server.
    \item The server fulfills these requests and resolves any resultant conflicts (\emph{e.g.} two agents attempting to occupy the same spatial location).
    \end{itemize}
\end{itemize}

The server proceeds in real-time and imposes noise on both the agents' observations and actions~\cite{noda2003robocup}. It is the responsibility of each agent to submit its action requests at the appropriate times to stay synchronised with the soccer server. Furthermore, each agent is allocated an individual CPU process, with no direct inter-process communication permitted. The soccer server provides a low-bandwidth, indirect communication method between agents by supporting simulated verbal commands.

\subsubsection{2D simulation league}

The 2D simulation league involves circular players being modeled with an $(x,y)$ position and orientation $\theta$. Each agent also maintains a head angle relative to its global orientation, allowing control of its field of view within human-like constraints. The action commands available to each agent include the following:
\begin{itemize}
\item Turn body or neck by a specified angle
\item Dash forward or backward with a specified ``power"
\item Slide tackle in a specified direction
\item Kick the ball in a specified direction with a specified angle, if near
\item Catch the ball if near (goalkeeper only)
\item Communicate with other players, either ``verbally" or by ``pointing" at a specified position
\end{itemize}

Each team consists of 11 players and a ``coach", which is a non-playing agent responsible for the allocation of players to each position given a number of randomly generated physical profiles (including characteristics such as speed and stamina). The 2D simulation league does model the dynamics or kinematics of any given human or robot; instead, it encourages development of complex player behaviours and team strategies~\cite{gliders2013tdp,gliders2014tdp,bai2012wrighteagle}. It is also a powerful framework for evaluating the emergent downstream effects (\emph{e.g.} team performance) of small perturbations to the systems underlying individual agents, as demonstrated in our recent study of particle filtering and self-localisation~\cite{budden2013particle}.

A screenshot from the 2D simulation league graphical client is presented in Figure~\ref{fig:screenshotsim}, and the 2D soccer server is available online at: \url{http://sourceforge.net/projects/sserver/}

\subsubsection{3D simulation league}

\begin{figure}[t]
\begin{center}
\includegraphics[width=13cm]{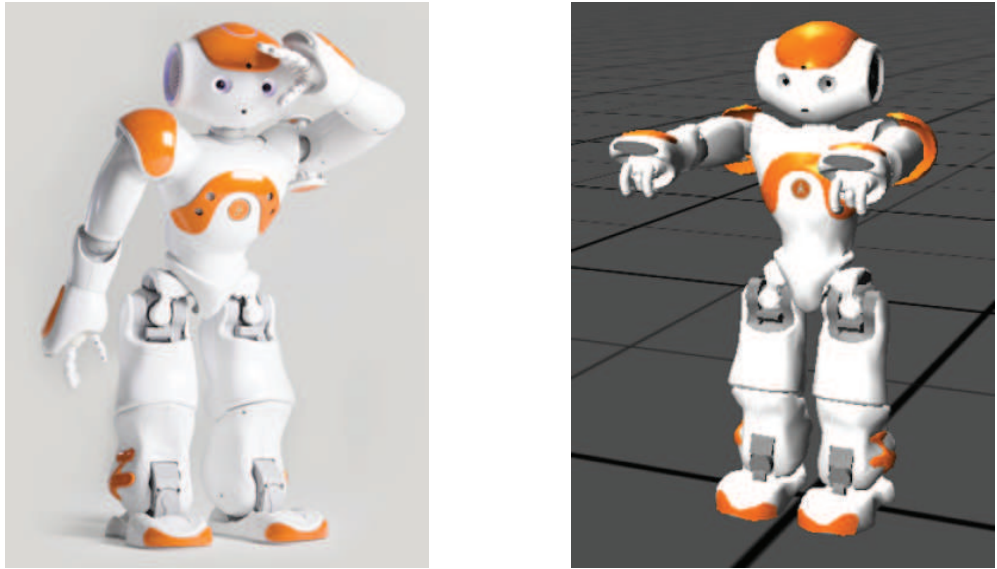}
\end{center}
\caption{The simulated NAO (right) is well-suited to global optimisation across the high-dimensional parameter spaces characteristic of bipedal locomotion systems, as experiments may be automated, parallelised and replicated exactly. This approach has proven very successful in identifying near-optimal parameter sets for subsequent local optimisation on the actual Aldebaran NAO (left)~\cite{macalpine2012design,niehaus2007gait}, highlighting the utility of simulation leagues for the investigation and improvement of physical robotic systems.\label{fig:nao}}
\end{figure}

The 3D simulation league implements a physically realistic world model and action interface, similar to that of physical robots rather than human players~\cite{bai2012wrighteagle}. Specifically, the simulator uses the Open Dynamics Engine library for simulation of rigid body dynamics, collision detection and friction, based on a model of the Aldebaran NAO humanoid robot (demonstrated in Fig.~\ref{fig:nao}). To remain consistent with the anatomy of the NAO, each agent simulates the following:
\begin{itemize}
\item 22-DOF in a 57 cm, 4.5 kg humanoid robot (six in each leg, four in each arm, two in the neck)
\item Perceptors, providing the agent with noiseless measurements of each joint position every simulation cycle
\item Effectors, allowing the agent to specify a direction and torque for each joint
\end{itemize}
Although no noise is introduced to perceptor and effector signals (with the exception of that resulting from approximations in the physics engine), the 3D simulation league introduces the non-trivial challenges of enabling each agent to stably walk, kick, dive and stand up from a fallen position. This makes it an ideal framework for global optimisation (and benchmarking optimisation algorithms) across the high-dimensional parameter spaces characteristic of bipedal locomotion systems. Although the simulated agents do not perfectly model the stochasticity inherent to actual NAOs, this approach has proven very successful in identifying near-optimal parameter sets for subsequent local optimisation on the physical robot itself~\cite{macalpine2012design,niehaus2007gait} (often in a lower-dimensional principle component space).

A screenshot from the 3D simulation league graphical client is presented in Figure~\ref{fig:screenshotsim}, and the 3D soccer server is available online at: \url{http://sourceforge.net/projects/simspark/}

\subsection{Enabling replicable and robust analysis}

Collectively, the simulation leagues provide ideal platforms for investigating emergent properties of complex robotic systems. Most team games and sports (both real and virtual) are characterised by rich, dynamic interactions that influence the contest outcome in a non-transitive manner. As described by Vilar \emph{et al.}, ``quantitative analysis is increasingly being used in team sports to better understand performance in these stylized, delineated, complex social systems"~\cite{vilar2013science}. Early examples of such quantitative analysis include \emph{sabermetrics}, which attempts to ``search for objective knowledge about baseball" by considering statistics of in-game activity~\cite{grabiner2004sabermetrics}. A recent study by Fewell \emph{et al.} involved the analysis of basketball games as networks, with properties including degree centrality, clustering, entropy and flow centrality calculating from measurements of ball position throughout the game~\cite{fewell2012basketball}. This idea was extended by Vilar \emph{et al.}, who considered the local dynamics of collective team behaviour to quantify how teams occupy sub-areas of the field as a function of ball position~\cite{vilar2013science}. Recently, Cliff \emph{et al.} presented several information-theoretic methods of quantifying dynamic interactions in soccer games, using the RoboCup 2D simulation league as an experimental platform~\cite{cliff2013towards}.

In addition to allowing high-level analysis of robotic systems overall, the simulation leagues provide inherent support for massively-parallel processing. This property has been leveraged for the development and analysis of algorithms with widespread applications in robotics; \emph{e.g.} optimising bipedal locomotion~\cite{macalpine2012design,niehaus2007gait}, self-localisation from noisy perception data~\cite{budden2013particle} and planning complex multi-agent strategies without direct agent-to-agent communication~\cite{gliders2013tdp,gliders2014tdp}. Although simulation league agents have only noisy perception of their environment, the soccer server itself has perfect information regarding the global state, enabling replicable quantification of experimental performance (\emph{e.g.} walk speed/stability and localisation accuracy).

\subsection{Wider implications in robotics research}

\begin{figure}[t]
\begin{center}
\includegraphics[width=14cm]{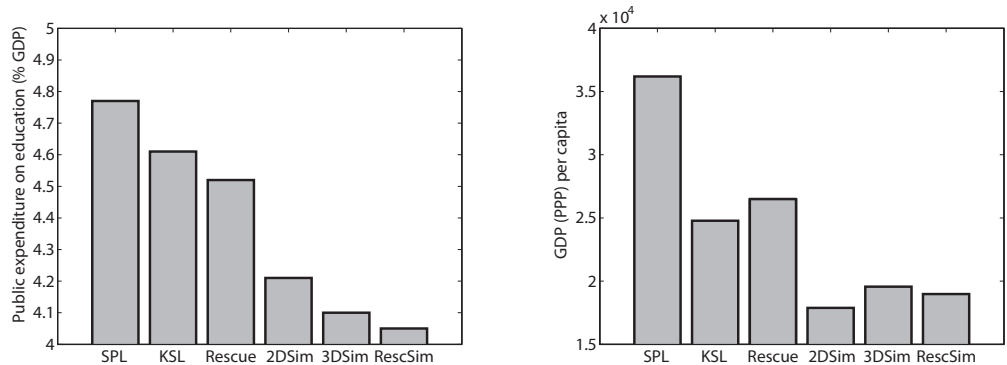}
\end{center}
\caption{PEoE (public expenditure on education as a percentage of GDP~\cite{world2012world}) and GDP/cap (gross domestic product at purchasing power parity per capita~\cite{world2012gdp}) for the home country of each participating RoboCup 2013 team, averaged over each of the six largest RoboCup leagues. Each of the three major simulation leagues (2D, 3D and simulated rescue) exhibit significantly lower values than those requiring the purchase or development of physical robots (standard platform league, kid-sized league and physical rescue). \label{fig:gdponed}}
\end{figure}

Robots are generally expensive to purchase, maintain and transport, creating an intractably-high entry barrier for institutes with less access to research funding. By abstracting away the requirement for physical robots, the RoboCup simulation leagues allow such institutes to actively contribute to many fields of robotics research. To validate this assertion, Fig.~\ref{fig:gdponed} presents the PEoE (public expenditure on education as a percentage of GDP~\cite{world2012world}) and GDP/cap (gross domestic product at purchasing power parity per capita~\cite{world2012gdp}) for the home country of each participating RoboCup 2013 team, averaged over each of the six largest RoboCup leagues.


\section{Simulation league case study: Analysis of competition formats}

The simulation league supports fully-automated, massively-parallel analysis of complex robotic systems, enabling replicable and robust investigation of algorithms and higher-level emergent behaviours. In the following sections, we leverage these properties to expand upon our previous analysis of RoboCup competition formats (\emph{e.g.} round robin) to determine which best approximates the true performance rankings of competing teams~\cite{rcsymp2014}.

The selection of an appropriate competition format is critical for both the success and credibility of any competition. Unfortunately, this choice is not straightforward: The format must minimise randomness relative to the true performance ranking of teams while keeping the number of games to a minimum, both to satisfy time constraints and retain the interest of participants and spectators alike. Furthermore, maintaining competition interest introduces a number of constraints to competition formats; \emph{e.g.} multiple games between the same two opponents (the obvious method of achieving a statistically significant ranking) should be avoided, making the resolution of non-transitive performance difficult.

\subsection{RoboCup competition formats}
\label{ss:formats}
The following competition formats were adopted to determine the final rank of the top 8 RoboCup 2D simulation league teams from 2012 to 2014:

\begin{itemize}
\item \textbf{2012:} The top 4 teams played 6 games each (3 quarterfinals (round robin), 2 semifinals and classification matches for 1st-versus-2nd and 3rd-versus-4th), and the bottom 4 teams played 4 games each. 
\item \textbf{2013:} A double-elimination system was adopted, where a team is ineligible for 1st place upon losing 2 games. 14 games were played in double-elimination format (\emph{i.e.} $2n-2, n=8$) followed by 2 classification games. 
\item \textbf{2014:} Our proposed format was adopted~\cite{rcsymp2014}. Specifically, a round-robin was conducted for the top 8 teams (28 games) followed by 4 classification games for 1st-versus-2nd, 3rd-versus-4th, etc.
\end{itemize}

Previously, it has been unclear whether these changes in competition format improve the fairness and reproducibility of the final team rankings. In general, lack of reproducibility is due to non-transitivity of team performance (a well-known phenomena that occurs frequently in actual human team sports).

\subsection{Methods of ranking team performance}
\label{sec:methodsofranking}

Before evaluating different competition formats, it is necessary to establish a fair (\emph{i.e.} statistically significant) ranking of the top 8 RoboCup 2D simulation league teams for previous years. This was accomplished by conducting an 8-team round-robin for previous years, where all 28 pairs of teams play approximately 1000 games against one another. In addition, two different schemes were considered for point calculation~\cite{rcsymp2014}:

\begin{itemize}
\item \textbf{Continuous scheme:} Teams are ranked by sum of average points obtained against each opponent across all 1000 games.
\item \textbf{Discrete scheme:} Firstly, the average score between each pair of teams (across all 1000 games) is rounded to the nearest integer (\emph{e.g.} ``1.9 : 1.2" is rounded to ``2 : 1"). Next, points are allocated for each pairing based on these rounded results: 3 for a win, 1 for a draw and 0 for a loss. Teams are then ranked by sum of these points received against each opponent.
\end{itemize}

The final rankings generated for 2012 and 2013 RoboCup 2D simulation league teams under these two schemes are presented in~\cite{rcsymp2014}. Although both schemes have been shown to generate statistically-robust results, we have chosen to adopt the continuous scheme for this study to avoid the boundary effects inherent to discretisation. In order to formally capture the overall difference between two rankings $\mathbf{r^a}$ and $\mathbf{r^b}$, the $L_1$ distance was utilised:

\begin{equation}
\label{eq:1}
    d_1(\mathbf{r^a}, \mathbf{r^b}) = \|\mathbf{r^a} - \mathbf{r^b}\|_1 = \sum_{i=1}^n |r^a_i-r^b_i| \ ,
\end{equation}

\noindent{where $i$ is the index of the $i$-th team in each ranking, $1 \leq i \leq 8$.}

\subsection{Results}
\label{ss:statsig}

Following iterated round-robin and continuous ranking scheme described in Sec.~\ref{sec:methodsofranking}, statistically significant rankings were generated for the top 8 RoboCup 2D simulation league teams for 2012, 2013 and 2014.  The full set of experiments is described in~\cite{rcsymp2014}, which verified that the proposed format (later adopted for RoboCup 2014) consistently outperformed the other candidates in terms of approximating the true rankings for RoboCup 2012 and 2013. Here, we expand upon this analysis to incorporate results for RoboCup 2014.

The $L_1$ distance (see eq. (\ref{eq:1})) was used to capture the discrepancy between RoboCup finals results, $\mathbf{r^a}$, and the statistically-significant rankings generated from the 28,000 game round-robin, $\mathbf{r^c}$:

\begin{align*}
d_1(\mathbf{r^a}, \mathbf{r^c})_{2012} &= 12\\
d_1(\mathbf{r^a}, \mathbf{r^c})_{2013} &= 12,
\end{align*}

\noindent{where the competition formats for RoboCup 2012 and 2013 are described in Section~\ref{ss:formats}. We can also quantify the corresponding discrepancy, $\mathbf{r^p}$, had our proposed competition format~\cite{rcsymp2014} been used for those competitions:}

\begin{align*}
d_1(\mathbf{r^p}, \mathbf{r^c})_{2012} &= 4\\
d_1(\mathbf{r^p}, \mathbf{r^c})_{2013} &= 6.
\end{align*}

\noindent{Our proposed format was subsequently adopted for the RoboCup 2014 finals, and the divergence from statistically-robust team rankings was equivalently small:}

\begin{equation*}
d_1(\mathbf{r^a}, \mathbf{r^c})_{2014} \Longleftrightarrow d_1(\mathbf{r^p}, \mathbf{r^c})_{2014} = 4.
\end{equation*}

\begin{figure}[t]
\begin{center}
\includegraphics[width=14cm]{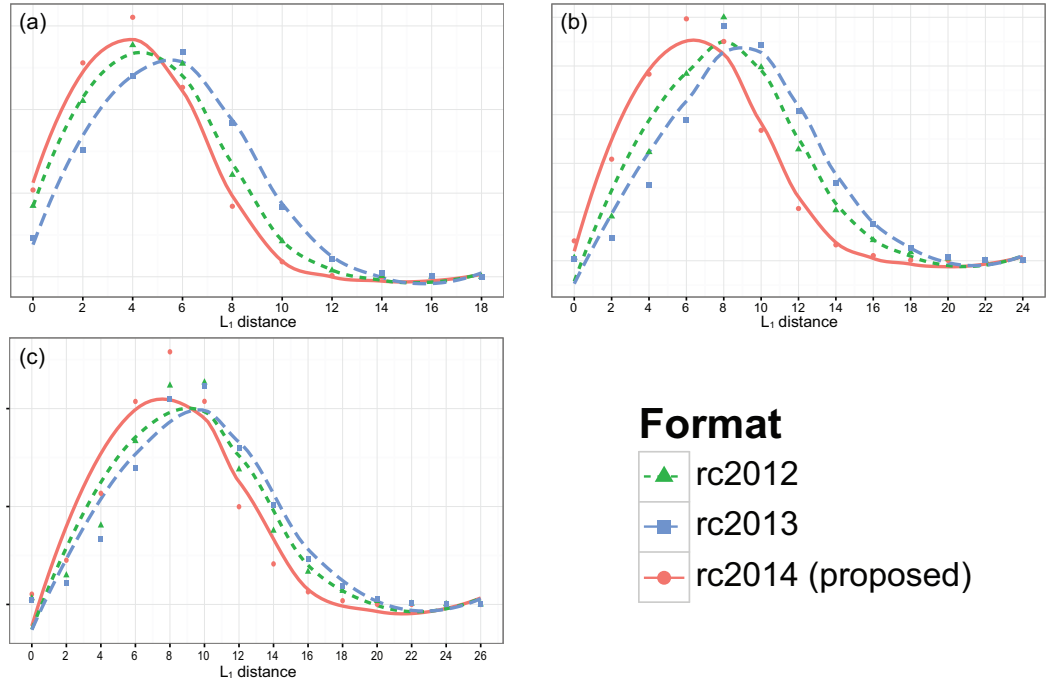}
\end{center}
\caption{Discrepancy between tournament and true team rankings, captured as an $L_1$ distance (see eq. (\ref{eq:1})), for 10,000 randomly-generated tournaments structured according to the three considered formats. It is evident that the proposed format (red) yields more statistically robust rankings (\emph{i.e.} smaller $L_1$ distance) than the formats adopted in RoboCup 2012 (green) and 2013 (blue), considering the top 8 teams from RoboCup 2012 (a), 2013 (b) and 2014 (c). \label{fig:comparison}}
\end{figure}

To statistically validate that the proposed competition format is significantly more appropriate than those adopted at RoboCup 2012 and 2013, 10,000 tournaments were generated for each format by randomly sampling game results from the 28,000 game round-robin. For each tournament, the $L_1$ distance, $d_1(\mathbf{r^a}, \mathbf{r^b})$ (see eq. (\ref{eq:1})), was calculated to capture the discrepancy between the tournament and true team rankings. These results are presented in Fig.~\ref{fig:comparison} for the top 8 teams from RoboCup 2012, 2013 and 2014. It is evident that the proposed format yields more statistically robust rankings (\emph{i.e.} smaller $L_1$ distance) than the formats adopted in previous years. Importantly, this analysis of competition formats would be impossible without the support for massively-parallel processing inherent to simulation leagues.

\section{Summary and discussion}

Continual increases in data volume and computational power have led to increased complexity in experimental methodologies across most fields of research. It is therefore unsurprising that many fields (particularly in the life sciences~\cite{hurley2014virtual}) have recently placed increased focus on enabling measurable, replicable and statistically-robust results. Although robotics researchers face many unique challenges due to the expense and stochasticity inherent to physical robots, we propose that physically-realistic simulated environments (epitomised by the RoboCup simulation leagues) have an important and widespread role to play in the future of robotics.

The simulation leagues often serve as platforms for the initial development and evaluation of software modules for later integration into physical robots~\cite{macalpine2012design,niehaus2007gait}, and many of these modules have applications well-beyond the RoboCup domain (\emph{e.g.} localisation and mapping~\cite{budden2013particle}). They also enable investigation of high-level emergent properties of complex robotic systems, as demonstrated in a recent study by Cliff \emph{et al.} that presents novel information-theoretic methods for quantifying dynamic interactions in a multi-agent context~\cite{cliff2013towards}.

In this study, we have provided an overview of the RoboCup simulation leagues (both 2D and 3D) and described their properties as they pertain to replicable and robust robotics research. To demonstrate their utility directly, we leverage the ability to run massively-parallelised experiments to evaluate different competition formats (\emph{e.g.} round robin) for the RoboCup 2D simulation league. Our results demonstrate that a hybrid format~\cite{rcsymp2014} minimises fluctuations from `true' (statistically-significant) team performance rankings within the time constraints of the RoboCup world finals.

Our experimental analysis and many others in previous literature~\cite{macalpine2012design,niehaus2007gait,budden2013particle,cliff2013towards} would be impossible with physical robots alone and have widespread applications beyond the scope of simulated soccer matches. We encourage other researchers to explore the potential for enriching their experimental pipelines with simulated components, both to minimise experimental costs and enable others to replicate and expand upon their results in a hardware-independent manner.

\bibliographystyle{splncs}
\bibliography{refs}

\end{document}